\documentclass[final]{l4dc2024}

\title[Continual Learning of Multi-modal Dynamics with External Memory]{Continual Learning of Multi-modal Dynamics with External Memory}
\usepackage{times}
\usepackage{enumitem}
\usepackage{tikz}
\usepackage{booktabs}
\usepackage{multirow}
\usepackage{wrapfig}
\usepackage{graphicx}
\usepackage{tikz}

\usetikzlibrary{bayesnet} 
\usetikzlibrary{backgrounds}
\usetikzlibrary{calc,quotes,arrows.meta}
\usetikzlibrary{shapes.misc,positioning,calc,graphs,arrows}

\pgfdeclarelayer{edgelayer}
\pgfdeclarelayer{nodelayer}
\pgfsetlayers{edgelayer,nodelayer,main}

\definecolor{hexcolor0xbfbfbf}{rgb}{0.749,0.749,0.749}

\tikzset{>=latex}
\tikzstyle{none}   = [inner sep=0pt]
\tikzstyle{line}   = [ -, thick, shorten <=1pt, shorten >=1pt ]
\tikzstyle{arrow}  = [ ->, thick, shorten <=1pt, shorten >=1pt ]
\tikzstyle{ardash} = [ dashed, ->, thick, shorten <=1pt, shorten >=1pt ]

\tikzstyle{box} = [rectangle, minimum width=1.5cm, minimum height=1.5cm,text centered, draw=black, inner sep=7pt]
\tikzstyle{neuron} = [circle, minimum width=4mm, very thick, draw=blue!80!black]

\tikzstyle{empty}=[circle,opacity=0.0,text opacity=1.0,inner sep=0pt]
\tikzstyle{box}=[rectangle,fill=White,draw=Black]
\tikzstyle{filled}=[circle,thick,fill=hexcolor0xbfbfbf,draw=Black]
\tikzstyle{hollow}=[circle,thick,fill=White,draw=Black]
\tikzstyle{param}=[rectangle,fill=Black,draw=Black,inner sep=0pt,minimum width=4pt,minimum height=4pt]
\tikzstyle{paramhollow}=[rectangle,thick,fill=White,draw=Black,inner sep=0pt,minimum
width=4pt,minimum height=4pt]

\usepackage{pgfplots}                               
\pgfplotsset{compat=newest}
\pgfplotsset{plot coordinates/math parser=false}
\usepgfplotslibrary{statistics}
\newlength\figureheight
\newlength\figurewidth
\newlength\figureheightsmall
\newlength\figurewidthsmall
\definecolor{POSTcolor}{rgb}{0.48, 0.20, 0.58} 
\definecolor{Qcolor}{rgb}{0.00, 0.53, 0.22} 

\author{%
 \Name{Abdullah Akgül} \Email{akgul@imada.sdu.dk}\\
 \addr University of Southern Denmark, Odense, Denmark
 \AND
 \Name{Gozde Unal} \Email{gozde.unal@itu.edu.tr} \\
 \addr Istanbul Technical University, Istanbul, Turkey
 \AND
 \Name{Melih Kandemir} \Email{kandemir@imada.sdu.dk}\\
 \addr University of Southern Denmark, Odense, Denmark%
}

\begin{document}

\maketitle

\begin{abstract}%
 We study the problem of fitting a model to a dynamical environment when new modes of behavior emerge sequentially. The learning model is aware when a new mode appears, but it cannot access the true modes of individual training sequences. The state-of-the-art continual learning approaches cannot handle this setup, because parameter transfer suffers from catastrophic interference and episodic memory design requires the knowledge of the ground-truth modes of sequences. We devise a novel continual learning method that overcomes both limitations by maintaining a \textit{descriptor} of the mode of an encountered sequence in a neural episodic memory. We employ a Dirichlet Process prior on the attention weights of the memory to foster efficient storage of the mode descriptors. Our method performs continual learning by transferring knowledge across tasks by retrieving the descriptors of similar modes of past tasks to the mode of a current sequence and feeding this descriptor into its transition kernel as control input. We observe the continual learning performance of our method to compare favorably to the mainstream parameter transfer approach.%
\end{abstract}


\section{Introduction}\label{sec:intro}

Continual Learning (CL) aims to develop a versatile model that is capable of solving multiple prediction tasks which are presented to the model one task at a time. The model is then expected to learn the latest task as accurately as possible while preserving its excellence at the previous ones. Performance drop caused by the newly learned task is called \textit{catastrophic forgetting}. CL is essential for developing intelligent agents that can adapt to new environmental conditions not encountered during training. For instance, an autonomous driving controller may improve its policy based on new experiences collected during its customer-side lifetime. 

There exists a solid body of work that adopts parameter transfer across tasks as the key element of task memorization \citep{kirkpatrick2017overcoming, nguyen2018variational, singh2019sequential, zenke2017synaptici}. There also appear preliminary studies on building attentive memories to capture tasks \citep{garnelo2018neural, fraccaro2018generative}. Principles that yield memory mechanisms optimal for CL are yet to be discovered. There has been prior work that focuses on CL for Recurrent Neural Networks (RNN) \citep{cossu2021continualrnn} but either classification or instance forecasting for time series. Complementarily, we study CL in the context of dynamical system identification. Probabilistic State-Space Models (SSM) are the gold standard methodology for the inference of complex latent dynamics. SSMs are widely applicable to forecasting impactful quantities such as weather, currency exchange, equity prices, and sales trends. SSM research gains significance also in robot learning parallel to the growing interest in model-based reinforcement learning \citep{hafner2019planet, hafner2020dreamer}.

\begin{wrapfigure}{r}{0.61\textwidth}
    \centering
    \includegraphics[width=\linewidth]{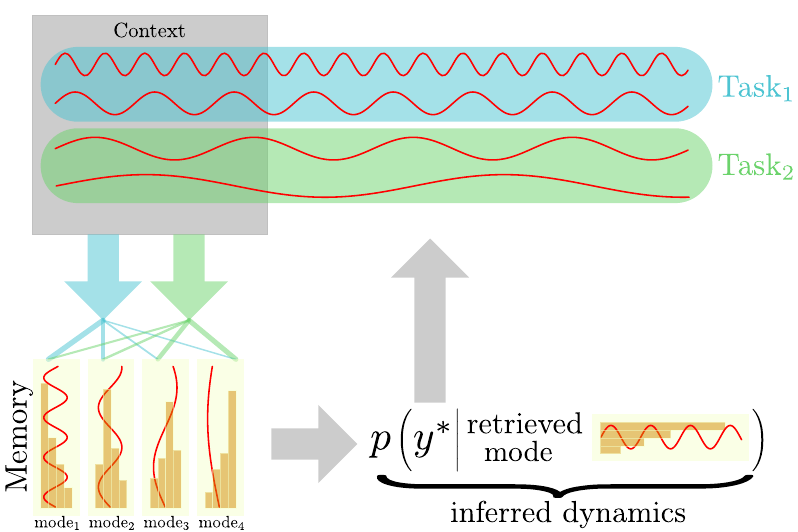}
    \caption{Our research shows that inferring and using mode descriptors from observed sequences significantly enhances CL performance. Storing these descriptors in external neural episodic memory and integrating them into subsequent tasks drives improved learning outcomes.}
    \label{fig:main-idea}
\end{wrapfigure}
Our main contributions are two-fold. First is a novel problem setup where dynamical system modeling tasks emerge sequentially and a probabilistic SSM is expected to learn them cumulatively. We assume each task to follow multi-modal dynamics, where each individual sequence of a task follows one of the possible modes that describe the task. Successful CL in such a setup presupposes maximally efficient encoding of tasks into long-term memory and their accurate retrieval.  Our second contribution is a novel CL model tailored for addressing this challenging problem. The state of the art in CL uses either parameter transfer or end-to-end differentiable attentive episodic memory for knowledge transfer across tasks. The multimodality of the sequences would undermine the parameter transfer approach due to catastrophic interference 
(the effect of global parameter updates on unintended regions of the input space) 
and it would require the ground-truth modality labels for the memory-based approaches to be applicable. Our approach sidesteps these limitations by capturing the characteristics of unknown modes of sequences of the present task into fixed-sized vectors, called \textit{mode descriptors}, and storing these descriptors in an external neural episodic memory addressable via a learnable attention mechanism. Our approach represents multiple task modes by feeding these mode descriptors into the state transition kernel as an additional input. We place a Dirichlet Process (DP) prior on the attention weights of the memory to encourage the explanation of the data with a minimum number of modes. Figure~\ref{fig:main-idea} illustrates our model with external memory and the problem with two tasks and four modes. Our resulting Bayesian model can be efficiently trained using a straightforward adaptation of existing variational SSM inference techniques.

We evaluate our model in two time series prediction data sets and three synthetic data sets generated from challenging nonlinear multi-modal dynamical systems. The performance of our method improves consistently over the established parameter transfer approach, verifying the importance of parsimonious use of neural episodic memory for efficient acquisition of knowledge within tasks and effective transfer of knowledge across different tasks.

\section{Continual Multi-Modal Dynamics Learning} \label{sec:problem_setup}
We assume a learning agent that observes a dynamical environment via sequences $y_{1:T} = \{y_1, \ldots,  y_T \}$ consisting of time-indexed measurements $y_t \in \mathcal{Y}$ living in a measurable space $\mathcal{Y}$. We denote a snippet of a sequence $y_{t:t'} = \{ y_t, \ldots, y_{t'} \}$ for an arbitrary time interval $[t,t']$. Modes refer to general distinguishable properties of a dynamic system such as states with different characteristics of a sine wave (amplitude, frequency) or semantically different dynamics such as different character trajectories. We define the \textit{mode} of a sequence $y_{1:T}$ as a fixed-sized vector and $m$ as an element of a $K-$dimensional embedding space $\mathcal{X}$. A dynamical system can potentially operate within a large number of modes. Making an analogy to the real world, an autonomous vehicle accounts for different environmental characteristics when planning and controlling for different weather conditions, countries, and times of a day. Only a few of the modes are active for a particular time point and each mode instantly activates and deactivates for limited time periods.

We search for a learning algorithm that enables the agent to fit a dynamical environment that has perpetually changing global characteristics. \emph{We cast the corresponding learning problem as CL of multi-modal dynamical systems, where each task is defined as a group of modes that modulate a specific dynamical system.} In formal terms, a task $\mathcal{T}_i$ is defined as a group of mode descriptors sampled from a mode generating oracle $P(\mathcal{X})$, that is 
\begin{align}
    \mathcal{T}_i &=  \{ m_r^{i} | m_r \sim P(\mathcal{X}), r = 1, \cdots, R \}, \quad i = 1, \cdots U, 
\end{align}
where $R$ is the number of modes per task and $U$ is the number of tasks. We assume that each sequence of a task follows dynamics modulated by a mode sampled from  $\mathcal{T}_i$
\begin{align}
    \mathcal{D}_{\mathcal{T}_i} = \Big \{y_{1:T}^n \Big | m_n & \sim P(\mathcal{T}_i),  y_{1:T}^n \sim p(y_{1:T}|m_n),  n = 1, \cdots, N \Big \},
\end{align}
where $N$ is number of instances per task, $P(\mathcal{T}_i)$ is a probability mass function defined on the modes of task $\mathcal{T}_i$, and $p(y_{1:T}|m_r)$ is the probability measure that describes the true behavior of the environment dynamics under mode $m_r$ within a time interval $[1, T]$. The marginal distribution of a task with respect to its modes is given as
$
  p(y_{1:T}|\mathcal{T}_i) = \sum_{r=1}^R p(y_{1:T}|m_r) P(m_r|\mathcal{T}_i).
$
The agent observes a task via a data set $\mathcal{D}_{\mathcal{T}_i}$ that contains only the sequences $y_{1:T}^n$ \textit{but not their corresponding modes}.

We are interested in learning a model that minimizes the true CL risk function below
\begin{align}
    \mathbb{R}_{\mathbb{L}}^{CL}(h_\theta) = \lim_{U \rightarrow +\infty} \sum_{i=1}^{U} \mathbb{E}_{\mathcal{T}_i \sim P(\mathcal{X})} \Big [\mathbb{E}_{y_{1:T} \sim p(y_{1:T}|\mathcal{T}_i)}  \left[ \mathbb{L}(h_{\theta}(\cdot|y_{1:C}), y_{C+1:T})  \right] \Big ], \label{eq:setup}
\end{align}
which amounts to the limit of the cumulative risk of individual tasks as they appear one at a time.
Above, $h_{\theta}: \mathcal{Y}^{C} \rightarrow \mathcal{Y}^{T-C}$ is a stochastic process that can map an observed sequence of an arbitrary length $C$ to the subsequent ${T-C}$ time steps. We call the first $C$ observations in the sequence $y_{1:C}$ as the \textit{context} and define $\mathbb{L}$ as a sequence-specific loss function defined on the future values of the sequence $y_{C+1:T}$. We evaluate the performance of predictions $\widehat{y}$ via two scores: \textit{Normalized Mean Squared Error (NMSE)} as a measure of the prediction accuracy of $h_{\theta}$ when used as a Gibbs predictor, and \textit{Negative Log-likelihood (NLL)} as a measure of Bayesian model fit that quantifies the model's own assessment on the uncertainty of its assumptions.
We approximate the true CL risk by its empirical counterpart:
$
    \widehat{\mathbb{R}}_{\mathbb{L}}^{CL} = \frac{1}{N} \sum_{i=1}^U \sum_{n=1}^{N} \mathbb{L} \Big( h_{\theta}(y_{C+1:T}^n|y_{1:C}^n), y_{C+1:T}^n \big |\mathcal{T}_i \Big)
$
for a finite number $U$ of tasks presented to $h_{\theta}$ one at a time.

\section{Novel Baseline: Variational Continual Learning for Bayesian State-Space Models}
\label{sec:bssm}

We build our target model on a Bayesian treatment of state-space modeling, which is proven to be effective in learning under high uncertainty and knowledge transfer across tasks, as practiced in the seminal prior art of CL \citep{kirkpatrick2017overcoming, nguyen2018variational}. We perform approximate inference using variational Bayes due to its multiple successful applications to state-space models \citep{frigola_gpssm2014, doerr2018prssm, ialongo2019overcoming} and its favorable computational properties. \textit{As there does not exist any prior work tailored specifically towards CL for dynamical systems, we curate our own baseline}. We adopt the established practice of setting the posterior of the learned parameters of the previous task as the prior of the next one and determine Variational Continual Learning (VCL) \citep{nguyen2018variational} as the state-of-the-art representative of this approach. Elastic Weight Consolidation (EWC) \citep{kirkpatrick2017overcoming} also follows the same approach but uses a simpler posterior inference scheme.

\paragraph{Bayesian State Space Models (BSSM):} are characterized by the data generating process below
\begin{align}
    \theta \sim p(\theta),  \qquad
    x_0  \sim p(x_0), \qquad \label{eq:bssm}
    x_t | x_{t-1}, \theta \sim p(x_t | x_{t-1}, \theta), \qquad
    y_t | x_t  \sim p(y_t | x_t),
\end{align}
where $x_t$ and $y_t$ correspond to the latent and observed state variables for time step $t$, respectively. The system dynamics are modeled by the first-order Markovian transition kernel $p(x_t | x_{t-1}, \theta)$ parameterized by $\theta$ that in turn follows a prior distribution $p(\theta)$. The latent states are mapped to the observation space via a probabilistic observation model $p(y_t | x_t)$. We formulate the initial latent state $x_0$ as another random variable that follows the prior distribution $p(x_0)$.

\paragraph{Variational Inference} is required to approximate the posterior $p(x_{0:T}, \theta | y_{1:T})$, which will be intractable for many choices of distribution families for the data generating process in Eq.~\ref{eq:bssm}. Following \cite{yildiz2019ode2vae}, we choose the variational distribution to be mean-field across the parameters of the dynamics and the latent states
\begin{align}
    q_{\xi, \psi}(x_{0:T}, \theta |y_{1:T}) &= q_{\xi}(x_0|y_{1:C}) \prod_{t=1}^T p(x_t | x_{t-1}, \theta) q_{\psi}(\theta),
\end{align}
where $q_{\xi, \psi}$ is an approximation to the true posterior $p(x_{0:T}, \theta | y_{1:T})$ with variational free parameters $(\xi, \psi)$. This formulation has multiple advantages. Firstly, modeling the marginal posterior on the initial latent state $x_0$ by amortizing on the context observations $y_{1:C}$ makes the Evidence Lower BOund (ELBO) calculation invariant to the context length. Secondly, adopting the prior transition kernel avoids duplicate learning of environment dynamics with twice as many free parameters and prevents training from instabilities caused by the inconsistencies between prior and posterior dynamics.  Applying Jensen’s inequality in a conventional way, the corresponding ELBO will be
\begin{align}
    &\log p(y_{1:T}) \geq \mathbb{E}_{p(x_{1:T}|\theta,x_0) q_{\xi}(x_0|y_{1:C}) q_\psi(\theta)} [ \log p(y_{1:T}|x_{1:T})] \label{eq:elbo_bssm} \\ 
    &\qquad\qquad\quad -KL(q_{\xi}(x_0|y_{1:C}) || p(x_0)) - KL(q_\psi(\theta) || p(\theta)), \nonumber
\end{align}
where $KL(\cdot||\cdot)$ stands for the Kullback-Leibler (KL) divergence between the two distributions on its arguments, and $X = \{x_0, \cdots, x_T \}$.

\paragraph{VCL for BSSMs} can be implemented as follows. Having fitted the ELBO (Eq.~\ref{eq:elbo_bssm}) on the data set for the first task which has observed sequences $\mathcal{D}_{\mathcal{T}_1}$, we attain $(\xi_1^*,\psi_1^*) = \mathrm{argmax}_{\xi, \psi} \mathcal{L}(\xi, \psi, \mathcal{D}_{\mathcal{T}_1})$.
When the next task arrives with data $\mathcal{D}_{\mathcal{T}_2}$, we assign $p(\theta) \leftarrow q_{\psi_1^*}(\theta)$ and maximize the ELBO again  $ (\xi_2^*,\psi_2^*) = \mathrm{argmax}_{\xi, \psi} \mathcal{L}(\xi, \psi, \mathcal{D}_{\mathcal{T}_2})$ .
We repeat this process continually for every new task. We refer to this newly curated baseline in the rest of the paper as \textit{VCL-BSSM}. We neglect the coreset extension of VCL since its application to BSSMs is tedious and its advantage is not demonstrated with sufficient significance in static prediction tasks studied in original work. 

\section{Target Model: The Continual Dynamic Dirichlet Process}
The commonplace Bayesian approach to CL transfers knowledge across tasks by assigning the learned posterior on the parameters of the previous task as the prior on the parameters of the current task. This is an effective approach when the subject of transfer is a feed-forward model, such as a classifier in a supervised learning setup \citep{nguyen2018variational} or a policy network in reinforcement learning \citep{kirkpatrick2017overcoming}. \textit{We conjecture that the existing parameter transfer Ansatz would not be sufficient for the transfer of more complex task properties such as modes of dynamical systems.} We address this problem by tailoring a novel CL approach from an original combination of an aged statistical machine learning tool, the DP, with modern neural episodic memory and attention mechanisms.

\paragraph{Dirichlet Processes} are stochastic processes defined on a countably infinite number of categorical outcomes, every finite subset of which follows a multinomial distribution drawn from a Dirichlet prior \citep{teh2006hdp}. A DP follows a Griffiths-Engen-McCloskey (GEM) distribution \citep{pitman2002combinatorial} 
\begin{align}
    \pi'_r | \alpha_0 \sim \mathrm{Beta}(1, \alpha_0), \qquad\qquad
    \pi_r = \pi'_r \prod_{j=1}^{r-1} (1-\pi'_j), \qquad\qquad
    \pi = [\pi_1, \ldots, \pi_R],
\end{align}
which we denote in short hand as $\pi \sim \mathrm{GEM}(\alpha_0, R)$. The GEM distribution can also be viewed as a stick-breaking process \citep{sethuraman1994dirichlet, fox2011sticky} where $\alpha_0>0$ is a scalar hyper-parameter. The data generation process below is called a DP for a base measure $G_0(\mathcal{X})$ defined on a $\sigma-$algebra $\mathbb{B}$  of $\mathcal{X}$
\begin{align}
        m_r \sim G_0(\mathcal{X}), \qquad\qquad\quad
        \pi \sim \mathrm{GEM}(\alpha_0, R), \qquad\qquad\quad
        G | \pi = \sum_{r=1}^{R} \pi_r \delta_{m_r}, 
\end{align}
where $\delta_{x}(A)$ is Dirac delta measure that takes value $1$ if $x \in A$ and $0$ otherwise for any measurable set $A \in \mathbb{B}$ and $G$ is a categorical distribution with parameters $\pi$. 

\paragraph{Neural Episodic Memory.} We assume that the environment dynamics can be expressed within a measurable latent embedding space $x_t \in \mathcal{X}$, and a sequence encoder $e_{\lambda}(x_{t:t'})$ parameterized by $\lambda$ for $t \leq t'$ that maps a sequence of latent embeddings into a fixed dimensional vector, as well as a neural memory $M = \{m_1, \cdots, m_R\}$ consisting of $R$ elements $m_r \in \mathcal{X}$ that live in the same space as latent embeddings $x_t$. We can construct a probability measure for $\mathbb{B}$ from the memory $M$ by updating
\begin{align}
    &m_r \leftarrow \label{eq:memory_update} (1-w_r(y_{{1:C}},m_r)) m_r +  w_r(y_{{1:C}},m_r) e_{\lambda}(y_{{1:C}})
\end{align}
for each sequence where 
$
    w_r(y_{{1:C}},m_r) = \dfrac{e^{\langle m_r,  e_{\lambda}(y_{{1:C}}) \rangle}} {\sum_{j=1}^R e^{\langle m_j,  e_{\lambda}(y_{{1:C}}) \rangle}}
$
for some similarity function $\langle \cdot, \cdot \rangle: \mathcal{X} \times \mathcal{X} \rightarrow \mathbb{R}^+$. This construction imposes a memory attention mechanism, where the encoded mode descriptors attend to the memory elements. Here we make the fair assumption that the modality of a sequence can be identified also from the observation space, while we need to infer the latent representations accurately to model the mode dynamics in detail. We choose an uninformative base measure that assigns equal prior probabilities to memory elements $G_0(M) = \sum_{k=1}^R \dfrac{1}{R} \delta_{m_r}$.

\paragraph{The full model.} We complement the BSSM in Eq. \ref{eq:bssm} with an external neural episodic memory $M$ that is updated for each observed sequence with the rule in Eq. \ref{eq:memory_update}. We place a DP prior on the retrieval of mode descriptors $m_r \in M$ to encourage the model to generate a minimum number of modes. We also feed the retrieved mode descriptor into the transition kernel $p(x_t|x_{t-1},m,\theta)$ as control input. The resultant model, which we call as the \textit{Continual Dynamic Dirichlet Process (CDDP)}, follows the generative process
\begin{align}
    \pi &\sim GEM(\alpha_0,R), \quad
    &m | \pi &\sim \sum_{r=1}^R \pi_r \delta_{m_r},\quad
    &\theta &\sim p(\theta), \\
    x_0 &\sim p(x_0), \quad
    &x_t|x_{t-1}, m, \theta &\sim p(x_t|x_{t-1}, m,  \theta), \quad
    &y_t|x_t &\sim p(y_t|x_t), \label{eq:dyn_ours} 
\end{align}
where memory capacity $R$ is set to a bigger number than the expected upper limit of the mode count. 

\paragraph{Inference.} Since $p(x_{0:T},\theta|y_{1:T})$ is intractable, we approximate it by variational inference. We inherit the advantages of the BSSM inference scheme by choosing the variational distribution as 
\begin{align}
    q_{\xi, \psi}(x_{0:T}, \theta, \pi |y_{1:T}) = q_{\xi}(x_0|y_{1:C}, \pi) \prod_{t=1}^T \sum_{r=1}^R q_\xi(\pi=r|y_{1:C}) p(x_t | x_{t-1}, m_r,  \theta) q_\psi(\theta), 
\end{align}
where
\begin{align}
    q_\psi(\theta) &= \mathcal{N}(\theta | \mu, \Sigma), \quad q_\xi(\pi |y_{1:C}) = Cat(w_1(m_1,y_{1:C}),\ldots,w_R(m_R,y_{1:C})), \\
    q_\xi(x_0|y_{1:C},\pi) &= \sum_{r=1}^R  \pi_r \mathcal{N}\Big (x_0 \Big | v_1( m_r, e_{\lambda}(y_{1:C})), v_2( m_r, e_{\lambda}(y_{1:C})) \Big)
\end{align}
with $x_0 \sim \mathcal{N}(0,I)$. In the expressions above, $v_1$ and $v_2$ refer to dense layers. The corresponding ELBO is then calculated with 
\begin{align}
    &\log p(y_{1:T}) \geq \sum_{r=1}^{R} q_\xi(\pi = r |y_{1:C}) \mathbb{E}_{p(x_{1:T}|\theta,x_0, m_r) q_\xi(x_0|y_{1:C}) q_\psi(\theta)} [ \log p(y_{1:T}|x_{1:T})] \\
    &\quad- KL(q_\xi(x_0|y_{1:C}) || p(x_0)) \nonumber - KL(q_\psi(\theta) || p(\theta)) - \sum_{r=1}^R w_r(m_r,y_{1:C}) \log (w_r(m_r,y_{1:C}) / \pi_r ) \nonumber. 
\end{align} 
\paragraph{Prediction. } Having trained the model on the latest task $\mathcal{T}_i$, its posterior predictive distribution for new sequence $Y^*$ and corresponding latent embeddings $X = \{ x_0, \cdots, x_T \}$ reads
\begin{align}
    &p(Y_{C+1:T}^* | \mathcal{D}_{\mathcal{T}_i}, Y_{1:C}^*) = \\
    &\qquad\qquad \sum_{r=1}^{R} q(\pi = r |Y_{1:C}^*) \int_{X,\theta} q_\xi(x_0| Y_{1:C}^*, \pi = r) q_\psi(\theta)   \Big[ \prod_{t=C+1}^T p(y_t^* | x_t) p(x_t | x_{t-1}, \theta, m_r) \Big]. \nonumber    
\end{align}

\section{Related Work}
\paragraph{State Space Models.}  There exists a considerable body of work on Bayesian versions of SSMs that employ Gaussian Process as transition kernels \citep{ialongo2019overcoming} and perform variational inference.  Another vein of work named Recurrent SSMs \citep{hafner2019planet, hafner2020dreamer} models the transition dynamics as RNN that map a state to the next time step deterministically while admitting a random state variable from the previous time step as input and feeding its output to the distribution of this variable at the subsequent time step.

\paragraph{Attention and Memory in Neural Nets.} Neural Turing Machines \citep{graves2014neural} are the first examples of attention-based neural episodic memory use with external updates. Attentive Neural Process \citep{kim2018attentive} employs an attention network to build a neural stochastic process that is consistent over observed predictions.  Evidential Turing Process \citep{kandemir2022evidential} maintains an external memory that learns to feed a Dirichlet prior on class distribution with informative concentration parameters inferred during minibatch training updates.

\paragraph{Continual Learning} is an instance of meta-learning \citep{finn2017maml} where new tasks are introduced one at a time and a base model is expected to learn the newest task without forgetting the previous ones. Early approaches to CL such as EWC \citep{kirkpatrick2017overcoming} transfer knowledge by the transfer of either deterministic parameters or their inferred distribution using Fisher information. 
VCL \citep{nguyen2018variational} improves on EWC with a more comprehensive inference scheme.
Generalized VCL \citep{loo2021generalized} maximizes the same ELBO as VCL but uses $\beta$-VAE to prevent the training instability caused by the dominance of the KL-divergence term. We do not use $\beta$-VAE since our setup does not have this problem. 

\cite{knoblauch2020optimalclperfectmemory} explains the superior performance of memorization-based CL algorithms based on experience replay, core set, and episodic memory \citep{shin2017deepgenerativereplay, lopez2017gem, luders2016continualNTM} over regularization-based algorithms \citep{kirkpatrick2017overcoming}.
For the first time, our CDDP studies probabilistic multi-modal dynamics in a CL setting by knowledge transfer via learned mode descriptors maintained in external memory.

\paragraph{Continual Learning with Episodic Memory. } \cite{rios2019memoryganforCL} builds a memory by maximum sample diversity in order to reduce catastrophic forgetting by remembering samples of previous tasks. In \cite{guo2020improved}, a memory keeps sample random examples from previous tasks and these samples are used in the training of the new tasks. There are other works \cite{lopez2017gem, chaudhry2018agem} that use memory for CL however, all of them build the memory with the help of data points provided with the ground truth but in our setup mode labels are not provided. Hence our work is not comparable to theirs. As neither of them studies dynamics modeling, their adaption to our setup is not straightforward.

\section{Experiments}
\label{sec:experiments}
We evaluate the performance of CDDP rigorously on five challenging applications of CL to time series forecasting. We provide a supplementary that includes a comprehensive PyTorch implementation of the entire experimental process, encompassing the studied models, the data generation process of the used synthetic environments, as well as the performance evaluation procedures.

\paragraph{\textbf Baseline Selection.} As we are the first to investigate CL for BSSM inference, there is no prior work we can take as a baseline without significant adaptation. We determine knowledge transfer across tasks by setting the parameter posterior of the previous task as the prior of the next as the state-of-the-art approach in  CL. We adapt VCL, the most established variant of this approach, to BSSMs in \S\ref{sec:bssm} as a baseline that can maximally challenge our CDDP. 
Baseline methods must be generative due to the nature of the dynamic modeling problem. Furthermore, as stated in the problem setup (\S\ref{sec:problem_setup}), the correct mode labels are not provided therefore baseline methods need to be unsupervised generative. No method meets these requirements but we adapt the VCL method.

\begin{wrapfigure}{r}{0.65\textwidth}
    \centering
    \includegraphics[width=0.95\linewidth]{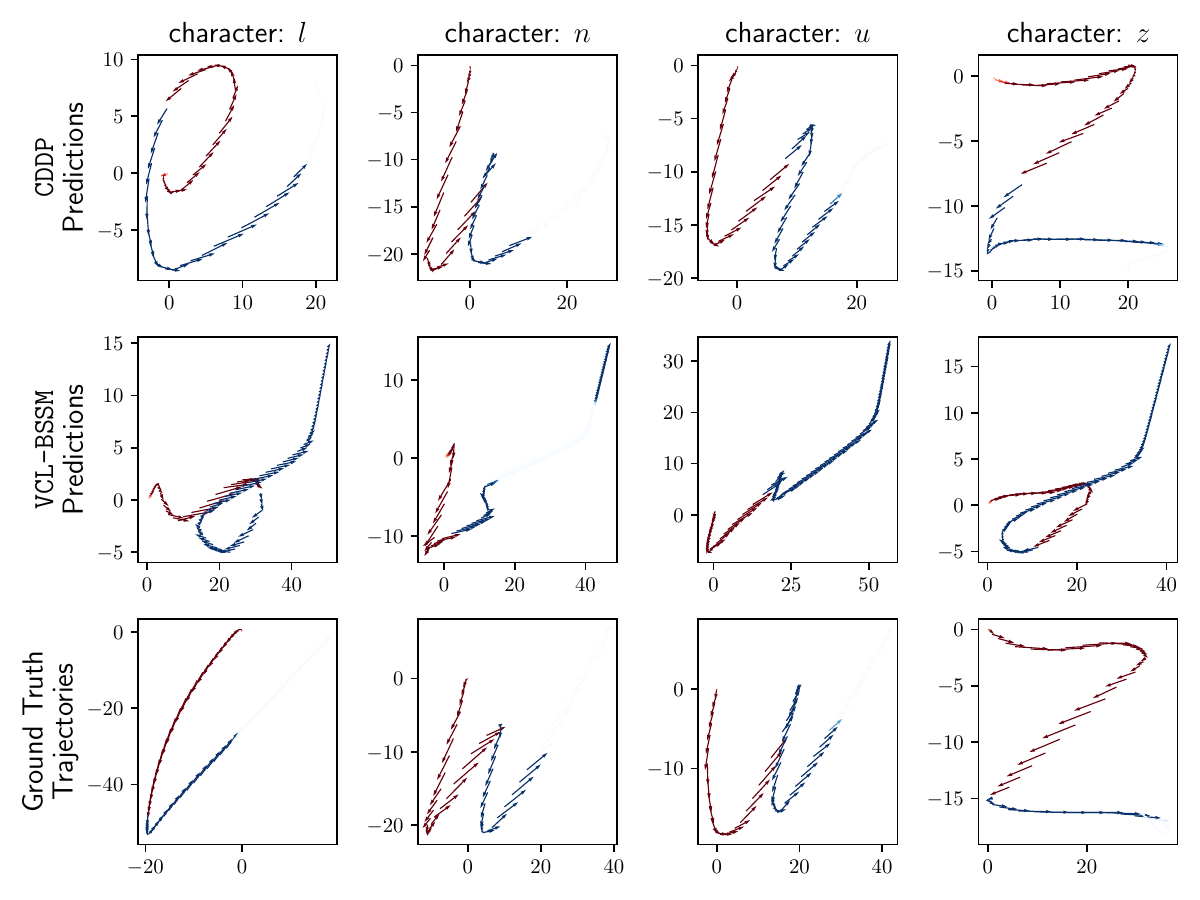}
    \caption{
    After learning is finished, CDDP and VCL-BSSM predictions on Character Trajectories are displayed alongside ground-truth trajectories. Context and predictions are plotted in red and blue respectively. CDDP exhibits superior memory of the initial task and precise adaptation to the last task due to its external episodic memory.
    }
    \label{fig:character_trajectories}
\end{wrapfigure}
\paragraph{Data Sets}
\paragraph{Synthetic Data Sets.}
We evaluate our CDDP on three nonlinear dynamical systems, where forecasting is challenging, while ground-truth task similarity is controllable. Too much task similarity would make CL unnecessary, while too much task difference would make it infeasible. When tasks share a reasonable degree of similarity, a successful CL algorithm is expected to capture, encode, and memorize these similarities and discard their differences. 

We generate modes that share similar dynamical properties, as their time evolution follows the same set of differential equations. However, modes differ from each other in the choice of the free parameters that govern the dynamical system. 
\textbf{i) Sine Waves} data set consists of signals grouped into modes described by different magnitudes and frequencies. Sine waves are created from the function $A \mathrm{sin}(2\pi f t)$ where $A$ is the magnitude, $f$ is the frequency, and $t$ is time. We generate different modes by changing the choices for magnitude and frequency levels.
For \textbf{ii) Lotka-Volterra} and \textbf{iii) Lorenz Attractor}, we follow the prior works \cite{haussmann2021pacbayes, satorras2019hybrid} and generate different modes by changing the free parameters of the systems. We perturb all three dynamical systems with Gaussian white noise for training splits. 

\begin{table}[t!]
    \centering
    \caption{Model performance on five datasets: Mean $\pm$ standard error over 10 repetitions. Area Under Curve (AUC) represents the averaged score across tasks.}
    \label{tab:results_table}
    \resizebox{0.99\linewidth}{!}{%
    \begin{tabular}{@{}p{1.4cm}ccccccccc@{}}
    \toprule
    \textbf{Data Set} &
    \textbf{\#modes} &
      \textbf{Model} &
      \multicolumn{1}{l}{\textbf{Score}} &
      \textbf{AUC} &
      \textbf{1 Task} &
      \textbf{2 Tasks} &
      \textbf{3 Tasks} &
      \textbf{4 Tasks} &
      \textbf{5 Tasks} \\ \midrule
    \multirow{4}{*}{\parbox{1.4cm}{Sine\\Waves}} &
    \multirow{4}{*}{15} &
      \multirow{2}{*}{\texttt{VCL-BSSM}} &
      NMSE &
      $1.00 \pm 0.04$ &
      $\mathbf{0.13 \pm 0.02}$ &
      $0.97 \pm 0.14$ &
      $1.22 \pm 0.15$ &
      $1.27 \pm 0.09$ &
      $1.39 \pm 0.08$ \\
     &
     &
       &
      NLL &
      $3.57 \pm 0.09$  &
      $\mathbf{ 1.76 \pm 0.22}$ &
      $3.75 \pm 0.24$ &
      $3.84 \pm 0.13$ &
      $4.20 \pm 0.15$ &
      $4.29 \pm 0.10$ \\ \cmidrule(l){3-10} 
     & 
     &
      \multirow{2}{*}{\texttt{CDDP}} &
      NMSE &
      $\mathbf{0.91 \pm 0.03}$ &
      $0.16 \pm 0.02$ &
      $\mathbf{0.88 \pm 0.11}$ &
      $\mathbf{1.07 \pm 0.15}$ &
      $\mathbf{1.12 \pm 0.14}$ &
      $\mathbf{1.31 \pm 0.11}$ \\ 
     &
     &
       &
      NLL &
      $\mathbf{3.50 \pm 0.09}$ &
      $2.09 \pm 0.27$ &
      $\mathbf{3.43 \pm 0.22}$ &
      $\mathbf{3.77 \pm 0.17}$ &
      $\mathbf{4.05 \pm 0.14}$ &
      $\mathbf{4.15 \pm 0.10}$ \\ \midrule
    \multirow{4}{*}{\parbox{1.4cm}{Lotka-Volterra}} &
    \multirow{4}{*}{8} &
      \multirow{2}{*}{\texttt{VCL-BSSM}} &
      NMSE &
      $\mathbf{0.58 \pm 0.04}$ &
      $0.17 \pm 0.03$ &
      $\mathbf{0.71 \pm 0.08}$ &
      $\mathbf{0.92 \pm 0.15}$ &
      $\mathbf{0.53 \pm 0.03}$ &
       \\
     &
     &
       &
      NLL &
      $1.50 \pm 0.05$  &
      $0.56 \pm 0.20$ &
      $1.97 \pm 0.28$ &
      $\mathbf{2.02 \pm 0.19}$ &
      $\mathbf{1.46 \pm 0.12}$ &
      \multicolumn{1}{l}{} \\ \cmidrule(l){3-9}  
     &
     &
      \multirow{2}{*}{\texttt{CDDP}} &
      NMSE &
      $0.60 \pm 0.06$     &
      $\mathbf{0.13 \pm 0.03}$ &
      $0.72 \pm 0.20$ &
      $0.96 \pm 0.18$ &
      $0.60 \pm 0.05$ &
       \\
     &
     &
       &
      NLL &
      $\mathbf{1.32 \pm 0.08}$ &
      $\mathbf{0.32 \pm 0.18}$ &
      $\mathbf{1.35 \pm 0.21}$ &
      $2.05 \pm 0.15$ &
      $1.54 \pm 0.21$ &
      \multicolumn{1}{l}{} \\ \midrule
    \multirow{4}{*}{\parbox{1.4cm}{Lorenz Attractor}} &
    \multirow{4}{*}{12} &
      \multirow{2}{*}{\texttt{VCL-BSSM}} &
      NMSE &
      $0.26 \pm 0.00$ &
      $0.22 \pm 0.02$ &
      $0.25 \pm 0.03$ &
      $0.27 \pm 0.02$ &
      $0.28 \pm 0.01$ &
       \\
     &
     &
       &
      NLL &
      $4.42 \pm 0.04$  &
      $4.22 \pm 0.16$ &
      $4.41 \pm 0.08$ &
      $4.55 \pm 0.09$ &
      $4.52 \pm 0.04$ &
      \multicolumn{1}{l}{} \\ \cmidrule(l){3-9}  
     &
     &
      \multirow{2}{*}{\texttt{CDDP}} &
      NMSE &
      $\mathbf{0.24 \pm 0.01}$ &
      $\mathbf{0.21 \pm 0.02}$ &  
      $\mathbf{0.24 \pm 0.02}$ &  
      $\mathbf{0.25 \pm 0.02}$ &  
      $\mathbf{0.26 \pm 0.02}$ &  
       \\
     &
     &
       &
      NLL &
      $\mathbf{4.35 \pm 0.06}$ &
      $\mathbf{4.06 \pm 0.19}$ &
      $\mathbf{4.33 \pm 0.02}$ &
      $\mathbf{4.53 \pm 0.06}$ &
      $\mathbf{4.46 \pm 0.03}$ &
      \multicolumn{1}{l}{} \\ \midrule
    \multirow{4}{*}{\parbox{1.4cm}{Libras}} &
    \multirow{4}{*}{15} &
      \multirow{2}{*}{\texttt{VCL-BSSM}} &
      NMSE &
      $\mathbf{0.14 \pm 0.00}$ &
      $\mathbf{0.13 \pm 0.01}$ &
      $0.14 \pm 0.00$ &
      $\mathbf{0.13 \pm 0.01}$ &
      $0.15 \pm 0.00$ &
      $\mathbf{0.14 \pm 0.00}$ \\
     &
     &
       &
      NLL &
      $-0.37 \pm 0.02$ &
      $-0.41 \pm 0.04$ &
      $-0.35 \pm 0.03$ &
      $\mathbf{-0.42 \pm 0.04}$ &
      $-0.32 \pm 0.03$ &
      $-0.36 \pm 0.03$ \\ \cmidrule(l){3-10} 
     &
     &
      \multirow{2}{*}{\texttt{CDDP}} &
      NMSE &
      $\mathbf{0.14 \pm 0.00}$ &
      $0.14 \pm 0.00$ &
      $\mathbf{0.13 \pm 0.01}$ &
      $0.14 \pm 0.01$ &
      $\mathbf{0.14 \pm 0.01}$ &
      $\mathbf{0.14 \pm 0.01}$ \\
     &
     &
       &
      NLL &
      $\mathbf{-0.39 \pm 0.04}$ &
      $\mathbf{-0.42 \pm 0.03}$ &
      $\mathbf{-0.40 \pm 0.04}$ &
      $-0.38 \pm 0.05$ &
      $\mathbf{-0.40 \pm 0.03}$ &
      $\mathbf{-0.37 \pm 0.04}$ \\ \midrule
    \multirow{4}{*}{\parbox{1.4cm}{Character\\Trajectories}} &
    \multirow{4}{*}{20} &
      \multirow{2}{*}{\texttt{VCL-BSSM}} &
      NMSE &
      $0.87 \pm 0.04$ &
      $\mathbf{0.42 \pm 0.04}$ & 
      $0.80 \pm 0.04$ &  
      $0.90 \pm 0.02$ &  
      $1.11 \pm 0.08$ &  
      $1.11 \pm 0.10$  \\
     &
     &
       &
      NLL &
      $0.14 \pm 0.02$  &
      $\mathbf{-0.23 \pm 0.06}$ &
      $0.06 \pm 0.05$ &
      $0.20 \pm 0.05$ &
      $0.31 \pm 0.08$ &
      $0.35 \pm 0.09$ \\\cmidrule(l){3-10} 
     &
     &
      \multirow{2}{*}{\texttt{CDDP}} &
      NMSE &
      $\mathbf{0.64 \pm 0.01}$ &
      $0.52 \pm 0.05$ & 
      $\mathbf{0.56 \pm 0.02}$ & 
      $\mathbf{0.69 \pm 0.04}$ & 
      $\mathbf{0.71 \pm 0.03}$ & 
      $\mathbf{0.71 \pm 0.02}$  \\ 
     &
     &
       &
      NLL &
      $\mathbf{-0.19 \pm 0.03}$ & 
      $-0.05 \pm 0.05$ &
      $\mathbf{-0.26 \pm 0.04}$ &
      $\mathbf{-0.17 \pm 0.03}$ &
      $\mathbf{-0.23 \pm 0.04}$ &
      $\mathbf{-0.26 \pm 0.04}$ \\ \bottomrule
    \end{tabular}%
    }
\end{table}

\paragraph{Real-World Data Sets.}
We evaluate our CDDP in two real-world time series classification data sets from \cite{dua2019UCI}. 
\textbf{i) Libras Movement Data Set} is the official Brazilian sign language and the data set consists of sequences of $(x,y)$ coordinates of hand movements from $15$ different classes.
\textbf{ii) Character Trajectories} data set consists of sequences of velocities of $(x,y)$ coordinates and pen force collected from hand-writings of $20$ English alphabet characters, which can be written by a single pen-down segment. Treating each class as a mode, we create multimodal time series forecasting tasks from these two data sets that satisfy the learning setup in Eq.~\ref{eq:setup}.

\paragraph{Context Length.} We select context length empirically as one-third of the sequence length amounting to five for Sine Waves, eight for Lotka-Volterra, 16 for the Lorenz Attractor, $15$ for Libras, and $35$ for Character Trajectories. The reason for that is, generally, one-third of the sequence captures the general characteristic of the dynamics and gives indications about modes.

\paragraph{Neural Network Architecture Details.} Our CDDP has four main architectural elements: 
\textbf{\textit{\underline{Encoder:}}} The sequence encoder $e_{\lambda}(x_{t:t'})$ governs the mean of our normal distributed recognition model $q_{\psi}(x_0|y_{1:C}, \pi)$. We feed $C$ observations as a stacked set of values into the encoder. The sequence encoder is a single dense layer for Sine Waves, Lotka-Volterra, Libras; and a multi-layer perceptron for the Lorenz Attractor, and Character Trajectories with two hidden layers. The perceptron uses the $tanh(\cdot)$ activation function followed by layer normalization.
\textbf{\textit{\underline{Decoder:}}} The likelihood function $p(y_t | x_t)$ of the base model serves as a probabilistic decoder that maps the latent state $x_t$ to the observed state $y_t$. We choose the emission distribution to be normal with mean governed by a single dense layer for Sine Waves, Lotka-Volterra, Libras; and a multilayer perceptron with two hidden layers for the rest. The perceptron uses the $tanh(\cdot)$ activation function followed by layer normalization.
\textbf{\textit{\underline{Transition Kernel:}}} We choose the transition kernel $p(x_t|x_{t-1},m,\theta)$ of CDDP to be a normal distribution with mean governed by a plain RNN that receives a concatenation of the previous hidden state and the mode descriptor as input. The RNN on the mean is a multilayer perceptron with one hidden layer. The perceptron uses the $tanh(\cdot)$ activation function followed by layer normalization. The transition kernel of the base model of VCL $p(x_t|x_{t-1},\theta)$ follows the same RNN architecture except that its input does not contain a mode descriptor.
\textbf{\textit{\underline{External Memory:}}} We set the memory size to $20$ for the Sine Wave environment, $10$ for Lotka-Volterra, $15$ for the Lorenz Attractor, $20$ for Libras, and $30$ for Character Trajectories.

\paragraph{Main Results.}  Table~\ref{tab:results_table} presents model performance throughout the whole CL process as the area under the learning curve. Our CDDP outperforms the parameter transfer-based VCL-BSSM baseline consistently in nearly all cases. Storing mode descriptors of the learned dynamics in an external memory, retrieving them in the subsequent tasks, and feeding them into the state transition kernel prevents catastrophic forgetting more effectively than plain parameter transfer. 
As seen in Figure \ref{fig:character_trajectories}, in the challenging character trajectories dataset, the prediction accuracy of our CDDP significantly outperforms VCL-BSSM in both the early and late stages of the CL period.
\begin{wraptable}{r}{0.65\linewidth}
\centering
\caption{Ablation study results given as the AUC on the Sine Waves data set. The reported numbers are mean $\pm$ standard error over $5$ repetitions.}
\label{tab:ablation}
\resizebox{\linewidth}{!}{%
\begin{tabular}{lccccc}
\toprule
\multicolumn{1}{l}{} &
  \textbf{\begin{tabular}[c]{@{}c@{}}Parameter  \\ Transfer\end{tabular}} &
  \textbf{\begin{tabular}[c]{@{}c@{}}Probabilistic\\ Parameters\end{tabular}} &
  \textbf{\begin{tabular}[c]{@{}c@{}}Memory\\ Content\end{tabular}} &
  \textbf{NMSE} &
  \textbf{NLL} \\ \hline
RNN                              & $\checkmark$ & $\times$    & N/A & $1.03 \pm 0.05$ & $3.50 \pm 0.10$ \\
VCL-BSSM                         & \checkmark   & \checkmark  & N/A & $0.93 \pm 0.03$ & $3.59 \pm 0.16$ \\ \hline
\multirow{4}{*}{CDDP Variants}   & $\times$     & \checkmark  & Zeros     & $0.94 \pm 0.04$ & $3.56 \pm 0.12$ \\
                                 & $\times$     & \checkmark  & Ones      & $1.12 \pm 0.13$ & $3.75 \pm 0.14$ \\
                                 & $\times$     & \checkmark  & Twos      & $1.35 \pm 0.23$ & $3.80 \pm 0.18$ \\
                                 & \checkmark   & \checkmark  & Learned   & $0.91 \pm 0.04$ & $3.56 \pm 0.12$ \\ \hline
CDDP Target                      & $\times$     & \checkmark  & Learned   & $\mathbf{0.87 \pm 0.03}$ & $\mathbf{3.35 \pm 0.15}$ \\ \hline
\end{tabular}%
}

\end{wraptable}

\paragraph{Ablation Study.} We investigate the contribution of individual design choices to the total performance of our target model. We study the effect of three design choices: i) knowledge transfer via parameters $\theta$ of the learned transition dynamics, ii) quantifying the uncertainty of the parameters of transition dynamics by a distribution $q_\psi(\theta)$, and iii) maintaining an external memory with learned or unlearned content. Table \ref{tab:ablation} shows a map of model variants corresponding to the activation status of these three design choices, as well as the corresponding numerical results on the Sine Waves data set over five repetitions. We observe a performance increase when knowledge transfer is done via the external memory \textit{instead of \textemdash but not together with \textemdash} parameter transfer, supporting the central assumptions of our target model. Setting the memory content to unlearned values causes rapid performance deterioration as values diverge from the learned values. This outcome demonstrates the essential role of the external memory in the CL performance of CDDP.

\section{Conclusion}
\label{sec:conc}
\paragraph{Summary. } We report the first study on CL of multi-modal dynamical systems. We curate a competitive baseline for this new problem setup from an adaptation of VCL to BSSMs. We introduce a novel alternative to the parameter transfer approach of VCL for within-task knowledge acquisition and cross-task knowledge transfer using an original combination of neural episodic memory, DPs, and BSSMs. We observe in CL of challenging multi-modal dynamics modeling environments that our alternative approach compares favorably to the established parameter transfer approach.

\paragraph{Broad Impact.} Our work can be used for varying applications such as: i) weather forecasting, where features can be transferred from one climate to another, ii) autonomous driving, where driving patterns can be adapted across different countries, and iii) model-based reinforcement learning algorithms, when the environment changes due either to the actions of the ego agent or to external factors. The memory architecture of CDDP may be improved by alternative embedding, update, and attention mechanisms. Our formulation of the transition rule is agnostic to the architecture that governs the transition kernel. 

\bibliography{main}

\end{document}